\documentclass{article}

\usepackage{authblk}
\usepackage{fullpage}

\usepackage[numbers,sort&compress,square]{natbib}
\usepackage{stackengine}

\usepackage[ruled,linesnumbered]{algorithm2e}
\usepackage{bbm}
\usepackage{natbib} 
\usepackage{amsmath,amssymb}
\usepackage{enumitem}
\usepackage{bm}
\usepackage{Definitions}
\usepackage{graphicx}
\usepackage{tikz}
\usepackage{subcaption}
\usepackage{caption}
\usepackage{xcolor}

\newcommand{\norm}[1]{\left\lVert#1\right\rVert}
\title{Pretrained Language Models are Symbolic Mathematics Solvers too!}

\author[1]{Kimia Noorbakhsh}
\author[2]{Modar Sulaiman\thanks{These authors contributed equally to this work.}}
\author[3]{Mahdi Sharifi$^{*}$}
\author[2]{Kallol Roy}
\author[3]{Pooyan Jamshidi}

\affil[1]{%
    Sharif University of Technology
}
\affil[2]{%
    University of Tartu
}
\affil[3]{%
    University of South Carolina
  }

\date{}

\begin{document}
\maketitle
\begin{abstract}
Solving symbolic mathematics has always been of in the arena of human ingenuity that needs compositional reasoning and recurrence. However, recent studies have shown that large-scale language models such as transformers are universal and surprisingly can be trained as  a sequence-to-sequence task to solve complex mathematical equations. These large transformer models need humongous amounts of training data to generalize to unseen symbolic mathematics problems. In this paper, we present a sample efficient way of solving the symbolic tasks by first pretraining the transformer model with language translation and then fine-tuning the pretrained transformer model to solve the  downstream task of symbolic mathematics. We achieve comparable accuracy on the integration task with our pretrained model while using around $1.5$ orders of magnitude less number of training samples with respect to the state-of-the-art deep learning for symbolic mathematics. The test accuracy on differential equation tasks is considerably lower comparing with integration as they need higher order recursions that are not present in language translations. We propose the  generalizability of our pretrained language model from Anna Karenina Principle (AKP). We pretrain our model with different pairs of language translations. Our results show language bias in solving symbolic mathematics tasks. Finally, we study the robustness of the fine-tuned model on symbolic math tasks against distribution shift, and our approach generalizes better in distribution shift scenarios for the function integration. \footnote{Preprint: Code and data are available at {\color{purple}\href{https://github.com/softsys4ai/differentiable-proving}{https://github.com/softsys4ai/differentiable-proving}.}}
\end{abstract}

\section{Introduction}
\label{Introduction}
Deep learning is a  ubiquitous choice in solving statistical pattern recognition problems of regression and classification. With a large training data set and compute power, they have proven to be very effective and achieve state-of-the-art performance in a wide range of tasks in natural language processing, computer vision, speech recognition, sentiment analysis, etc \citep{lu2021pretrained}. Though deep learning triumphs in the statistical domain \citep{bengio2003neural}, there is an active interest in extending deep networks in symbolic computation \citep{Symbolic, davis2019use, allamanis2017learning, zaremba2014learning, loos2017deep}. There are mainly two motivations for this: (i) performing symbolic mathematical tasks, such as symbolic integration and solving differential equations, in deep net architectures, and (ii) applying neural networks in the domain of automated theorem proving, computer algebra systems, and natural language understanding (NLU) that requires a symbolic knowledge system. The key capability of symbolic computation is that symbols maintain their identity as they do multiple roles, while deep neural networks exploit shared representation and composition. 

This paper uses a pretrained language model to solve symbolic mathematics tasks, particularly symbolic integration and differential equations. We show our pretrained transformer architecture on language translation is expressive enough to solve large class symbolic mathematics such as function integration and differential equations, which have traditionally been approached using logic and exhaustive search. Moreover, our pretrained model is \emph{sample efficient} and compute efficient--i.e., it requires fewer epochs to converge to good accuracy. The first major work of solving symbolic mathematics with transformer architecture is by \citet{Symbolic}. They use the transformer model that is mainly used for NLP tasks to solve the symbolic computation. They first re-frame the mathematical equations as text sequences and then solve those equations as a sequence-to-sequence translation. Their transformer model catches pattern in the mathematical expressions, e.g., the expressions of the form $\sin^{-1}(x)$ will have its primitive as $\frac{1}{\sqrt{1-x^{2}}}$. We extend the work of \citet{Symbolic} and train their symbolic math dataset by fine-tuning pretrained translation models to solve the downstream task of symbolic mathematics.
 The pretrained language model will transfer the syntactic and semantic structure of the present in the language, mathematical expressions represented as trees. The inherent limitation between the many-to-one map between mathematical expression and tree encoding is partially regularized by the pre-training with the language translation. For example, the same mathematical expressions $7 + 3 \times (5 + 2)$ and $3 \times (5 + 2) + 7 $ are encoded as different trees. We regularize (penalize) this freedom of expression of encoding a mathematical expression by multiple trees by pretraining our transformer model with language translation. The sentence in a language has an order as specified by the famous quote by J. R. Firt “You shall know a word by the company it keeps.”. Unlike language, where the meaning of a word is given by its neighbors, the value of a mathematical sub-expression (mathematical word)  is not influenced by its neighboring expressions. In \citet{Symbolic}'s training data set generation for function integration,  mathematical expressions $F$ and $G$ are  generated and the corresponding derivatives $f$ and $g$ are computed. The training dataset are the tuples $(f, F)$, $(g, G)$ and a new integration function dataset $Fg$ is generated (assuming $ (fG, \int fG)$ is in the training set) through IBP (Integration By Parts) \footnote{More details about the datasets are explained in Section \ref{datasets}.} method as: 

\begin{equation*}
 \int Fg = FG - \int fG.
\end{equation*}

Their vanilla transformer model during training learns to build the correlation between$\int Fg$ and $fG$ for solving symbolic mathematics. We differ from  their model by (i) forcing our transformer model to develop conditional probability between randomly generated functions $P_{\Theta}(f|G)$ and $P_{\Theta}(g|F)$ as follows: 
\begin{align*} 
P(fG)   &=  P(f|G) P(G) \\ 
P(Fg)   &=  P(g|F) P(F)
\end{align*}
where $P_{\Theta}$ is our pretrained transformer model and $\Theta$ is the learned parameter (weights and biases).  By re-framing the problem to a conditional probability model, we bypass the distributions of randomly generated functions $P(F)$ and $P(G)$. Our method also shows marginal robustness to different types of dataset generation method, as shown in table \ref{tab:gen2}. (ii) Our model’s predictions improve even when there is a difference of length between input and output sequence.  This is because of the phenomena of heavy-tailed distribution, where the model can generate rare small or large output expressions \citep{sornette2006critical, martin2018implicit}. 

The paper is organized as follows: In Section \ref{problem-formulation} we discuss the prediction of our pretrained transformer model in the language conditional probability and optimization, Section \ref{sec:theory} discusses our proposed theory and hypothesis which will be verified in our experiments, Section~\ref{Methodology}
discusses experimental setting and methodology, architecture, datasets, and the evaluation metric, and Section~\ref{Empirical} poses the following research questions and answers them:
\begin{enumerate}
    \item Does this pretrained model help us to use less data for fine-tuning?
    \item Does the result of this fine-tuning depend on the languages used for pretraining?
    \item How robust is this fine-tuned model with respect to the distribution shift of test data compared to fine-tuning data?
\end{enumerate}
Section~\ref{Related work and Discussion}, discusses literature review, and finally, Section~\ref{conclution} concludes the paper.

\section{Problem Formulation}
\label{problem-formulation}
Mathematical expressions can be depicted as binary-unary trees, with operators in the form of internal nodes, operands in the form of children, and numbers, constants, and variables in the form of leaves \citep{Symbolic}. These trees can be transformed into a sequence of math letters by traversing them in some specific order. In this paper, a tree of symbolic mathematical expressions is scanned by the prefix traversal to produce a sequence corresponding to the mathematical expression. We formulate our symbolic mathematics as a Seq2Seq translation problem with a large scale pretrained mBART (and Marian-MT) transformer. The pretrained transformer is fine-tuned with random expressions dataset for the case of function integration and differential equation. 

We model our mBART Transformer doing sequence-to-sequence translation from  the source  to the target language  as an Encoder–Decoder framework~\cite{DBLP:journals/corr/BahdanauCB14}. Though, the mBART Transformer
is distinct from the usual Encoder-Decoder framework of RNN (LSTM).  The burden to encode the input sequence now lies on the encoder with the trick of self-attention.
The encoder now tracks associations from its own input sequence, before passing it to the decoder. In a way, self-attention is looking inwards where the Encoder looks for the clues to  optimally encode its own sentence and has the following components: (i) \textbf{Key: Value} to label elements  in the sequence and its associated value (ii) \textbf{Query}: query the keys and select the best match for the request. We model our pretrained mBART Transformer as a probabilistic model, translating from the input mathematical sequence $\textbf{x} = ( x_{1}, \cdots , x_{\text{T}})$ to output mathematical sequence $\textbf{y} = ( y_{1}, \cdots , y_{\text{T}})$,

\begin{equation}
P(\textbf{y}) =  \Pi_{t=1}^{\text{T}} p(y_{t}| \{ x_{1}, \cdots , x_{t-1} \}, W )
\end{equation}
where $W$ is the pretrained weights of the transformer.

\section{Theory}
\label{sec:theory}
We propose a novel method of generalizability of our mBART Transformer model from evolutionary game theory and population dynamics. We group neurons that share a similarity as populations. In our mBArt transformer, we have two neuron groups $W_{\text{Pretraining}}$ and $W_{\text{Finetuning}}$. $W_{\text{Pretraining}}$ neuron population contributes to language attributes: grammar, syntax, and semantics. $W_{\text{Finetuning}}$ contributes to mathematical structures; binary and unary tree representations, and the height of the tree that encodes the mathematical sequence. Therefore, the attributes of $W_{\text{Pretraining}}$ and $W_{\text{Finetuning}}$ may seem quite disjoint initially and have little overlap. In our paper and experiments, we show the counterintuitive notion, that these two do overlap. Thus, pretraining with language data helps to solve symbolic mathematics problems. We propose that $W_{\text{Pretraining}}$ and $W_{\text{Finetuning}}$ are in evolutionary pressure and trying to survive. Here, the concept of survival is represented as the number density. For example, if the  number of members in population $\norm{W_{\text{Pretraining}}}$ is less than a threshold, we say $W_{\text{Pretraining}}$ population fails to survive. Partitioning the neurons of our model $W = [W_{\text{Pretraining}} | W_{\text{Finetuning}} ]$,  we model population dynamics of neurons with celebrated  predator-prey equations of Lotka–Volterra: 
\begin{equation}\label{Lotka–Volterra}
\begin{aligned}
\dot{\norm{W_{\text{Pretraining}}}} &=  \norm{W_{\text{Pretraining}}}(a - b\norm{W_{\text{Finetuning}}})  \\
\dot{\norm{W_{\text{Finetuning}}}} &=  \norm{W_{\text{Finetuning}}}(-c + d\norm{W_{\text{Pretraining}}}) 
\end{aligned}
\end{equation}

where $\dot{\norm{W_{\text{Pretraining}}}} = \frac{d\norm{W_{\text{Pretraining}}}}{dt}$, $\dot{\norm{W_{\text{Finetuning}}}} = \frac{d\norm{W_{\text{Finetuning}}}}{dt}$ are the time derivatives and constants  $a, b, c, d \geq 0$. The mapping between neuron population and its learning features capacity is a complex problem and not well understood. While most research efforts theorize, that models learn features hierarchically through gradient descent by updating the weights, we advocate a parallel population dynamics is running  as in Equation~\ref{Lotka–Volterra}. The features of a model learn is through the intertwining of stochastic gradient descent and population dynamics. Thus, there exists a differentiable map $T$ that maps from population  to feature set of the language sequences (e.g., number of unique words, average sequence length). For simplicity, we take a time dependent feature set  $F = \{f_{1}(t), f_{2}(t) \}$ that emerges from the only population dynamics. These features are dependent on the population of $||W_{\text{Pretraining}}||$ and $||W_{\text{Finetuning}}||$ as a linear combination. Solving time-independent Lotka–Volterra Equation\ref{Lotka–Volterra} we and plugging in Equation~\ref{feature} we get:

 \begin{gather}\label{feature}
 \begin{bmatrix} f_{1}(t)  \\ f_{2}(t)  \end{bmatrix}
 =
  \begin{bmatrix}
   T_{11}    &
   T_{12}    \\
   T_{21}    &
   T_{22}    
   \end{bmatrix}
\begin{bmatrix} \norm{W_{\text{Pretraining}}(0)}e^{at}  \\ \norm{W_{\text{Finetuning}}(0)}e^{-ct}  \end{bmatrix}
\end{gather}

where the 0 index represent population at time zero. From Equation~\ref{feature} we infer the feature vectors emerged from the population dynamics of $||W_{\text{Pretraining}}||, ||W_{\text{Finetuning}}||$. Therefore, survivability of neuron populations is projected as the emergence of features. This we call as \textit{feature fight} that emerges also from predator-prey equations of Lotka–Volterra equations. Some interesting concepts can be inferred from our proposed hypothesis:
\begin{enumerate}
  \item The features that emerge from interactions $||W_{\text{Pretraining}}||$ and $||W_{\text{Finetuning}}||$ are time-dependent~\cite{west2003developmental}.
  \item Different models trained in similar environments will behave similarly, and is called the Anna Karenina Principle (AKP)~\cite{wiki:Anna, diamond98,osti_10218957}.
\end{enumerate}
We verify (AKP) principle in our experiments as our mBart models which were pretrained in multiple translations' environment (English-Romanian, English-Greek, etc.) have similar generalizability (section \ref{language}). We argue that our pretrained transformer models' loss landscapes for different pretrained translations are similar~\cite{NEURIPS2018_a41b3bb3, szegedy2015rethinking, DBLP:journals/corr/abs-2106-07682}. Thus, the bottleneck of solving non-convex optimization during fine-tuning has a dependence on the evolutionary principles. Our $\norm{W_{\text{\text{Finetuning}}}}$ measure forces the transformer models to forget about the linguistics part of them learned during the fine-tuning. Our mathematical sequences have no inherent grammatical structure, and mBart needs to forget its grammar for generalizing to out-of-distribution (OOD) mathematical sequences. $\norm{W_{\text{Finetuning}}}$ is exactly doing as catastrophic forgetting and knocking off parts of $\norm{W_{\text{Pretraining}}}$ neurons which remember the grammar. Searching for parts of the neurons that remembers grammar is a difficult problem as described in Lottery Ticket Hypothesis~\cite{DBLP:journals/corr/abs-1803-03635}. Evolutionary game theory and population dynamics come to the rescue to solve the generalizability on mathematical sequences. In  summary, pretraining helps generalization.

\section{Experimental Setting}
\label{Methodology}
We evaluate a diverse set of symbolic mathematical data sets as introduced in \citet{Symbolic}. The tasks studied in these datasets include symbolic function integration and solving differential equations of order one and two. Mainly, we are interested in whether pretrained language models are genetically capable of solving these tasks with fewer data. Moreover, whether the language that they have been pretrained on impacts their result after transfer learning. In Section \ref{Empirical}, we will do this empirical study by asking structured research questions.

\subsection{Architecture}
\label{Architecture}
We use the Marian model \citep{mariannmt} and the mBART model \citep{mbart}, pre-trained on different translation tasks by the NLP group at the University of Helsinki and Facebook, using the Marian model and the mBART model of the famous NLP framework, Hugging-Face \citep{hugging-face}.

Both models follow the famous transformer architecture introduced in \citet{vaswani2017attention}. The Hugging-Face mBART model has an embedding size of $1024$, with $16$ attention heads and $12$ layers. The Marian-MT model we use (only) in Section \ref{language}, has an embedding size of $512$, with $8$ attention heads and $6$ layers. The Marian Model and the mBART model have approximately $74$ and $610$ a million parameters. The Parameter counts may vary depending on the vocab size of the language they have been pretrained on. We also train the model used in \citet{Symbolic} with the same parameters as the mBart model (i.e., with an embedding size of $1024$, $12$ layers and $16$ attention heads.).

\subsection{Datasets}
\label{datasets}
Thanks to \cite{Symbolic}, there is a good dataset resource for Symbolic Mathematics available publicly. In all the experiments in this paper, we use the same datasets as \citet{Symbolic}, or generate new datasets using the same generation methods.

For the mathematical integration task, there are three generation methods. Forward (FWD), Backward (BWD), and Integration by Parts (IBP). The forward approach, generates random functions and calculates their integrals with an external symbolic mathematical framework. The backward approach, on the other hand, generates a random function and then computes its derivative and adds the pair to the dataset with a backward manner. Both backward and forward approaches have some issues. The forward approach is only capable of creating samples that can only be integrated by a mathematical framework, and also the samples generated by this approach have short problems with long solutions. The backward approach normally generates samples that the integral is shorter than the equation itself. In contrast to the other two methods, the IBP approach uses the integration by parts to generate samples that do not need an external computer algebra framework, but in terms of the equation lengths, it is similar to the FWD approach (generates short problems and long solutions.) \citep{Symbolic}. The datasets for the first order differential equations are referred as ODE1 and the second order differential equations are referred as ODE2. Detailed information about datasets can be found at \citet{Symbolic}.

\subsection{Metric}
\label{metrics}
In all of our experiments, we report the Accuracy (similarly to \citet{Symbolic}), which is calculated by the accuracy of our predictions by comparing the generated equation and the reference equation. The generated equation by the models might not be in the same format as the reference equation; therefore, we simplify the difference between the predicted and the reference equation to check whether it is $0$ or not. It is also necessary to mention that all the results in section \ref{Empirical} are reported with the evaluations of beam size $1$. 

\section{Experimental Evaluation}
\label{Empirical}
In this section, we examine the results showing transfer from language translation to solving symbolic math equations and attempt to understand better why this happens and which factors enable this transfer. The following subsections include our research questions, how we design experiments to answer them, the discussions of the results, and their implications. Note that we refer to \citet{Symbolic}'s model results with the keyword \textbf{LC} in our tables and visualizations. 

We train our models with the Adam optimizer \citep{adam}, with a learning rate of $10^{-4}$. We run all of our experiments with the mBART and the Marian-MT model only for 15 epochs, while we train the LC model as long as the model converges (which usually takes around $100$ epochs.). \footnote{The experiments with the mBART model were performed on a machine equipped with one RTX A6000 NVIDIA GPU and 48 GB memory. The experiments with the Marian-MT model were performed on a machine equipped with one NVIDIA Tesla V100 GPU and 512 GB memory.}

\subsection{Does this pretrained model help us to use less data for training?}
\label{data}

\begin{figure*}[]
	\begin{subfigure}[b]{0.19\textwidth}
	    \centering
		\includegraphics[width=1\linewidth]{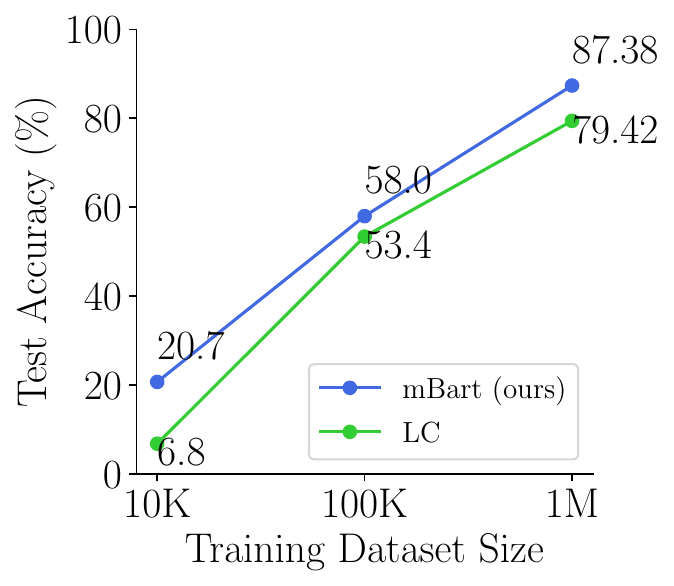}
		\caption{FWD}
	\end{subfigure}%
    \begin{subfigure}[b]{0.19\textwidth}
    	 \centering
		\includegraphics[width=1\linewidth]{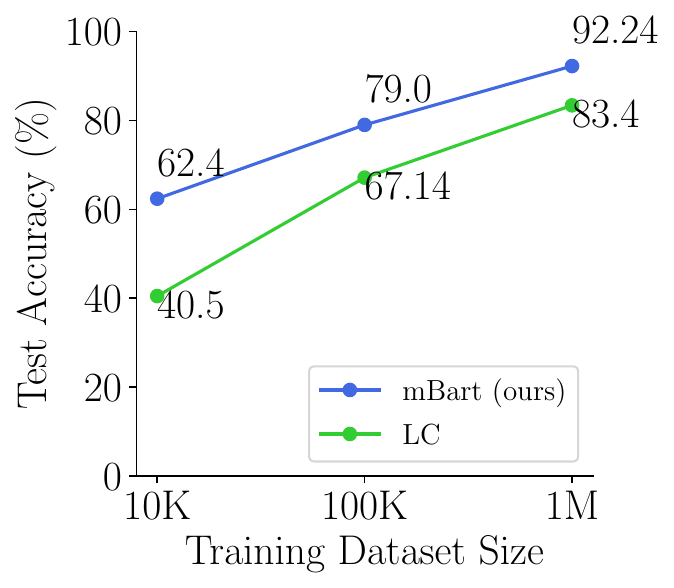}
		\caption{BWD}
	\end{subfigure}%
	\begin{subfigure}[b]{0.19\textwidth}
	   \centering
		\includegraphics[width=1\linewidth]{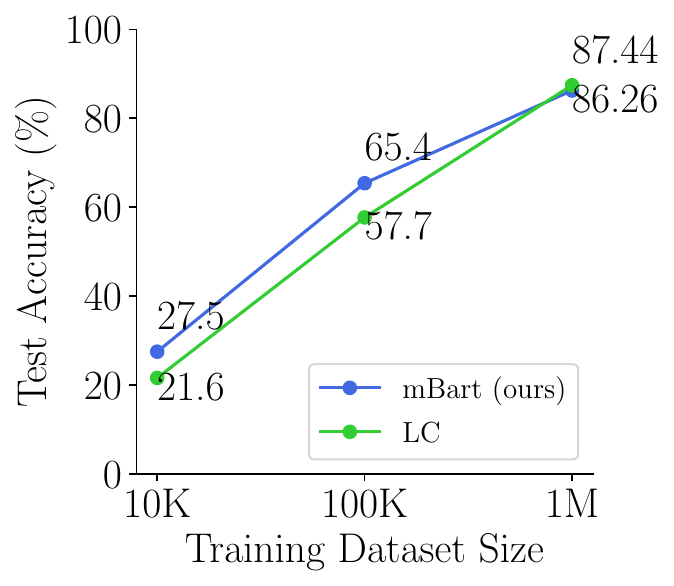}
		\caption{IBP}
	\end{subfigure}%
	\begin{subfigure}[b]{0.19\textwidth}
	\centering
		\includegraphics[width=1\linewidth]{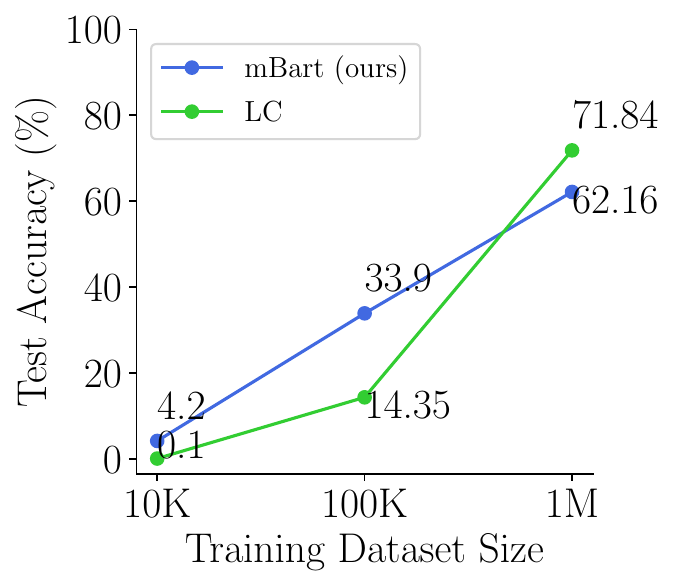}
		\caption{ODE1}
	\end{subfigure}
	\begin{subfigure}[b]{0.19\textwidth}
		\centering
		\includegraphics[width=1\linewidth]{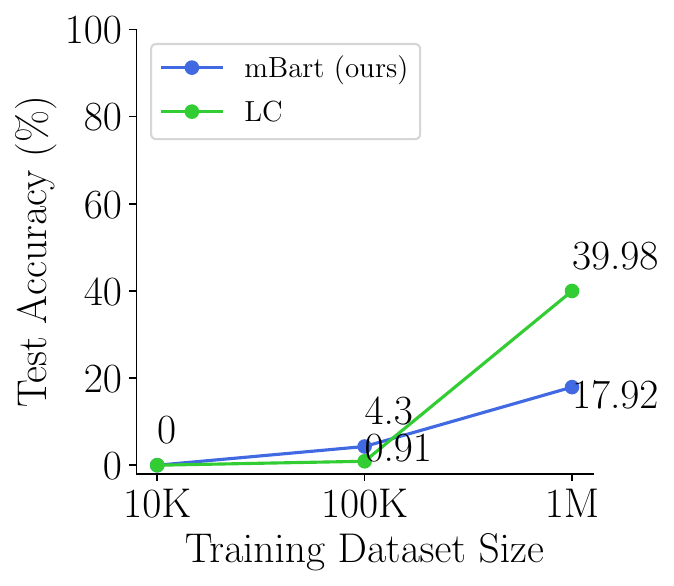}
		\caption{ODE2}
	\end{subfigure}
	
	\caption{The accuracy of our mBART language model and the LC model when trained on different training sample sizes. Panels (a), (b), and (c) are for the integration task.}
	\label{fig:accuracies}
\end{figure*}

\begin{table*}[]
\centering
\begin{tabular}{cccccc}
\hline
 & Integration (FWD) & Integration (BWD) & Integration (IBP) & ODE $1$   & ODE $2$   \\      \\ \hline
Our Model    & $87.4$         & $92.2$         & $86.2$        & $62.2$       & $17.9$  \\
LC's Model   & $79.4$         & $83.4$          & $87.4$        & $71.8$       & $39.9$  \\ \hline
\end{tabular}%
\caption{\small Accuracy of our models (in percentage ($\%$)) and the \citet{Symbolic}'s model on integration and differential equation solving. The number of training samples used to train the models in all tasks is 1 million (Both our model and LC model.). Results are tested on test data sets of size $5000$ samples.}
\label{tab:accuracy_table}
\end{table*}

As studied in \citet{Symbolic}, to train transformer architecture on the symbolic math data, we need a vast amount of training data for each task to achieve the highest accuracies (in the order of 40 million to 80 million training samples for each task.). We investigate if fine-tuning the pretrained models on language translation tasks on the symbolic math data can help us use considerably fewer data in the fine-tuning stage. 

In this section, we will use the pretrained mBART \citep{mbart} model for the English to Romanian translation task \footnote{The pretrained mBART model is available at {\color{purple}\href{https://huggingface.co/facebook/mBART-large-en-ro}{https://huggingface.co/facebook/mBART-large-en-ro}.}}, and fine-tune it on our math data (see Section \ref{datasets}). We report the accuracy of our models on the integration and differential equation solving in table \ref{tab:accuracy_table}. In this table, we use the same training dataset for both our mBART model and the LC model. We train our mBART model only for $15$ epochs for all $5$ tasks (FWD, BWD, IBP, ODE1, and ODE2), but we continue the training of the LC model until convergence (which takes around $100$ epochs for each task.). We can see in the table \ref{tab:accuracy_table} that our model outperformed in the integration task, with a considerable gap from the LC model. But it cannot properly perform on the differential equation task, especially the second-order differential equations.  

We extend this exploration by running the same experiment for different orders of magnitude of training data (i.e., 10K, 100K, and 1M). We report the test accuracy (see Section \ref{metrics}) of each experiment for both models (mBART and LC) in figure \ref{fig:accuracies}. Our model has higher accuracy in comparison to LC in all tasks and with different training sample sizes, except that in the differential equations the accuracy growth of our model suddenly gets lower than the LC model when using the 1 million samples for training. 

We achieve comparable accuracy on the integration task with our pretrained model while using around 1.5 orders of magnitude less number of training samples than the state-of-the-art model in \citet{Symbolic} (i.e, we use 1 million training samples against the 40-80 million samples that \citet{Symbolic} used for training their model.). As we have discussed previously in the Section \ref{sec:theory}, the mBART language model has already been pretrained by the language translation. During this pretraining, our mBART model searches for that hypothesis that outputs the shortest translated sequence (the shortest Romanian sequence for a given input of English sequence).  During the fine-tuning, it uses the same hypothesis learned previously to search for mathematical expressions that has minimum length. Also, because our mBART language model is very large, it is doing an internal look-up and search for the  solutions depth-wise in the mathematical expression tree. The model is thus effectively searching greedily  than the LC model. Note that the accuracies reported for the LC model in table \ref{tab:accuracy_table}, as well as in tables \ref{tab:gen1} and \ref{tab:gen2} are by training this model also with 1 million training samples (the high accuracies (over $95\%$) reported in \citet{Symbolic} are achieved by sample sizes of range $20-40$ million training samples).
 
\subsection{Are the results of such fine-tuning, language dependent?}
\label{language}

\begin{table*}[]
\centering
\begin{tabular}{c|ccccc}
\hline
Language           & Integration (FWD) & Integration (BWD)  & Integration (IBP) & ODE $1$ & ODE $2$ \\
\hline
English - Romanian &    $38.8$       &         $67.8$          &      $51.5$  &   $\bm{23.4}$    &   $1.8$      \\ 
\cline{1-6} 
English - Greek    &    $39.3$       &         $69.5$          &       $48.6$            &    $17.3$     &   $2.5$      \\ 
\cline{1-6}
English - Arabic   &    $43.9$       &       $71.3$     &       $\bm{53.5}$   &   $16.4$      &    $2.7$     \\ 
\cline{1-6}
English - French   &       $47.7$    &       $\bm{71.4}$            &      $52.5$             &    $18.9$     &   $2.9$      \\ 
\cline{1-6} 
English - Spanish  &         $43.5$  &       $70.4$            &     $51.8$              &  $18.7$       &     $\bm{3.3}$   \\ 
\cline{1-6} 
Greek - English    &         $39.1$  &        $69.1$           &        $47.9$           &    $16.2$     &    $2.2$     \\ 
\cline{1-6}
Arabic - English   &        $43.3$   &         $69.3$         &       $50.7$  &   $22.5$      &  $2.3$       \\ 
\cline{1-6}
French - English   & $\bm{50.5}$ &       $71.2$           &   $52.7$     &    $19.7$     &     $2.3$    \\ 
\cline{1-6} 
Spanish - English  &        $40.4$   &        $69.9$           &      $51.7$            &    $20.2$    &   $2.0$      \\ 
\hline
\end{tabular}%
\caption{Evaluation of accuracy of our Marian-MT model (in percentage ($\%$)) on the integration and differential equation solving for different pretrained languages. The highest accuracy is indicated by bold case in each column (task). We see that the language has no specific impact on the results of this fine-tuning.}
\label{tab:language_table}
\end{table*}
\begin{table*}[]
\centering
\begin{tabular}{c|ccccc}
\hline 
Evaluation Method           & Integration (FWD) & Integration (BWD)  & Integration (IBP) \\
\hline 
Ensemble  &        $65.6$   &         $83.5$            &     $74.9$             \\ 
\hline
Majority Voting  &        $49.1$   &         $72.4$           &     $59.4$           \\ 
\hline
\end{tabular}%
\caption{Ensemble-based evaluation of the combination of 9 Marian-MT models of table \ref{tab:language_table} (in percentage ($\%$)) on the integration task.}
\label{tab:ensemble}
\end{table*}

We investigate whether different languages used to train our pretrained models impact the results of this transfer learning. We wish to see whether the quality of the results in section \ref{data} might have been dependent on the specific source-target language in our language model, i.e., the learned representations. In other words, the specific language could have been a confounder. Therefore, to remove this confounds, we fine-tune our symbolic math data on 9 different pretrained language translation tasks containing various source-target languages.

To be able to perform more experiments on multiple languages (due to the computational costs), we fix our training sample size to 100K samples per task, and we use the pretrained Marian-MT model of Hugging-Face \citep{hugging-face} which has already been pretrained on many language translation tasks, and is available online \footnote{The pretrained Marian-MT models are available at {\color{purple}\href{https://huggingface.co/Helsinki-NLP}{https://huggingface.co/Helsinki-NLP}.}}. Since the accuracy of the models based on what we saw in Section \ref{data} are consistent, we only report the accuracies for the 100K sample dataset. Accuracies will not be optimal, but they are sufficient to answer our question. We test all the experiments on test datasets of size $1000$. The results are shown in table \ref{tab:language_table}. As we can see in this table, for each task, a different pretrained language has the highest accuracy (indicated in bold case.). For example, in the FWD task the French to English model had the highest accuracy and so on. Therefore, table \ref{tab:language_table} shows that the results of this fine-tuning approach are not language dependent and our hypothesis that language is a confounder for our results is not true.  

In addition to the above results, we conducted two other experiments to further investigate the impact of the pretraining dataset. These two experiments are both an ensemble-besed evaluation and combine the 9 models of \ref{tab:language_table} in their evaluation. First, we test the combination of the 9 pretrained models. To evaluate the accuracy in this scenario, we input each data sample to each of these 9 fine-tuned models and if at least one of them could successfully solve the symbolic math task, we consider our evaluation a success and if none of them was not able to solve the task accurately, we consider it as a failure. In the second evaluation we use majority voting to evaluate the combination of the 9 models, and we consider our evaluation successful, if and only if at least 5 of the 9 models could solve the task accurately. The results for the integration task are shown in table \ref{tab:ensemble}. \footnote{We only performed this experiment on the integration task because of it's higher accuracies and the fact that it gives us a more realistic intuition of the ensemble evaluation.} As we can learn from the table, the majority voting approach does not improve the overall accuracy compared to the ones in table \ref{tab:language_table}, whereas the ensemble approach significantly improves the results. This again confirms that there are no language dependencies in this approach.

It is also important to note that this Marian-MT model has an embedding size of $512$, which is twice less than the mBART model (and the LC model) we use in Section \ref{data}. But because our goal in this section is to study the impact of languages and there are many pretrained models available of Marian-MT, we choose to use this model in our language study.\footnote{Investigating the effect of embedding size more systematically to the results is considered as future work.}
\begin{table*}[htb]
    \small
    \centering
        \begin{tabular}{l|cc|cc|cc}\hline
            \multicolumn{1}{c|}{} & \multicolumn{2}{c|}{Forward} & \multicolumn{2}{c|}{Backward} & \multicolumn{2}{c}{Integration by parts} \\ 
            \multicolumn{1}{c|}{Training data} & Ours(mBART) & LC & Ours(mBART) & LC & Ours(mBART) & LC \\
            \hline 
            FWD                & $87.38$ & $79.42$ & $7.30$ & $6.90$ & $74.20$ &  $74.10$ \\ 
            BWD                & $12.82$&  $9.28$ & $92.24$ & $83.40$ & $24.02$ & $17.60$ \\ 
            IBP             & $30.46$ & $28.70$ & $35.00$ & $20.50$ & $86.26$ & $87.44$\\ \hline
        \end{tabular}
    \caption{\small Accuracy of the models (in percentage ($\%$)) on function integration. Results are tested on test data sets of size $5000$ samples. The models are trained on the 1 million sample size training data, as discussed in Section \ref{data}.}
    \label{tab:gen1}
\end{table*}
\begin{table*}[htb]
\centering
\begin{tabular}{c|c|lllll}
\hline
Testset Type &
  Metrics &
  \multicolumn{1}{c}{Integration (FWD)} &
  \multicolumn{1}{c}{Integration (BWD)} &
  \multicolumn{1}{c}{Integration (IBP)} &
  \multicolumn{1}{c}{ODE1} &
  \multicolumn{1}{c}{ODE2} \\ 
 \hline
 &
   Ours &
  \multicolumn{1}{c}{$60.6$} & 
  \multicolumn{1}{c}{$67.8$} &
  \multicolumn{1}{c}{$70.7$} &
  \multicolumn{1}{c}{$39.1$} &
  \multicolumn{1}{c}{$8.9$} \\ 
Polynomials & LC & 
 \multicolumn{1}{c}{$54.7$} & 
  \multicolumn{1}{c}{$60.0$} &
  \multicolumn{1}{c}{$80.1$} &
  \multicolumn{1}{c}{$60.6$} &
  \multicolumn{1}{c}{$57.9$} 
   \\
 
   \hline
\multicolumn{1}{l|}{} &
   Ours &
  \multicolumn{1}{c}{$91.9$} &
  \multicolumn{1}{c}{$87.0$} &
  \multicolumn{1}{c}{$78.9$} &
  \multicolumn{1}{c}{$48.3$} &
  \multicolumn{1}{c}{$10.6$} \\
Trigonometric & LC & 
 \multicolumn{1}{c}{$92.4$} & 
  \multicolumn{1}{c}{$85.8$} &
  \multicolumn{1}{c}{$91.8$} &
  \multicolumn{1}{c}{$74.4$} &
  \multicolumn{1}{c}{$60.6$} 
\\
  \hline
\multicolumn{1}{l|}{} &
  Ours &
  \multicolumn{1}{c}{$90.9$} &
  \multicolumn{1}{c}{$75.1$} &
  \multicolumn{1}{c}{$72.4$} &
  \multicolumn{1}{c}{$35.9$} &
  \multicolumn{1}{c}{$6.8$} \\
Logarithmic & LC & 
 \multicolumn{1}{c}{$87.9$} & 
  \multicolumn{1}{c}{$73.3$} &
  \multicolumn{1}{c}{$87.96$} &
  \multicolumn{1}{c}{$75.6$} &
  \multicolumn{1}{c}{$72.0$} 
\\
    \hline
\end{tabular}%
\caption{\small Accuracy of our models (in percentage ($\%$)) on the integration and differential equation solving for different pretrained languages. Results are reported on test datasets of different types (polynomial, trigonometric and logarithmic.), and of size $5000$.}
\label{tab:gen2}
\end{table*}
\subsection{How robust is this fine-tuned model with the distribution shift?}
\label{shift}
As also studied in \citet{Symbolic}, it is important to see whether these transformer models are biased towards the distribution of their training data or not. In order to evaluate this concept, we define two different kinds of distribution shift as follows:
\begin{itemize}
    \item The first one is only for the integration task and is similar to the Section $4.7$ in \citet{Symbolic}. Meaning that we will investigate how robust our models trained in \ref{data} are when we change their testing distribution. We report the evaluation metrics trained and tested on a different combination of training datasets in table \ref{tab:gen1}.
    \item The second kind of distribution shift that we are interested in is due to the modality of the test dataset. This type of distribution shift was not studied by \citet{Symbolic} and is a new type of distribution shifts we introduce in this paper. Each training sample we use on all tasks (in Sections \ref{data}, and \ref{language}) has a combination of all different types of equations such as polynomial, trigonometric, and logarithmic expressions. We want to see whether a model trained on this type of dataset can generalize to solve type-dominant functions (i.e, functions containing only polynomial equations or containing only trigonometric equations and so on.). Therefore, we generate different types of test data, varying in the kind of equation they represent, such as trigonometric equations, polynomial equations, and logarithmic equations. We test the ability of our models trained in \ref{data} to see which kinds of equations they can solve better, helping us to understand the impact of linguistic data better. The results are reported in table \ref{tab:gen2}.
\end{itemize}

Table \ref{tab:gen1} indicates that our mBART model is more robust with respect to the generation distribution shift (i.e., FWD, BWD and IBP method for integration task.) and can achieve comparable performance in comparison to the pure transformer model (LC) model. 

To evaluate the robustness of our approach in terms of different equation types, we created three different test datasets for each task. The first dataset is polynomial dominant, meaning that the samples of dataset were created mostly by polynomials without using trigonometric and logarithmic functions. The second and third datasets are trigonometric dominant and logarithmic dominant, respectively. This means that the trigonometric dominant dataset was created using mostly trigonometric functions, and the logarithmic dataset was generated using mostly logarithm and exponential functions. Table \ref{tab:gen2} indicates that our mBART model is not able to generalize to type dominant equations as well as the LC model can (except in the FWD and BWD approaches of the integration task.). The highest accuracies of both models are in their generalization to solve trigonometric expressions, and the lowest results are in pure polynomial ones. This agrees with our theory (see Section \ref{sec:theory}), because the mBART model tries to find the shortest sequence and the higher order polynomial equations are less compressible. Also, higher order polynomials need accurate precision (F64) for their representation.  On the other hand, trigonometric and the logarithmic equations can be compressed into shorter expressions (for example, $sin^2(x) + cos^2(x)$ is $1$. or $e^{ix}= \cos{x} + i\sin{x} $), and ;therefore, the performance on these two sets of type-dominant test samples are better.

\section{Related work and Discussion}
\label{Related work and Discussion}
\subsection{Transformers in different modalities}
Attention \citep{bahdanau2014neural} is a powerful mechanism led to recent achievements in developing strong DNN models in NLP like the transformer architecture \citep{vaswani2017attention}. Attention mechanism has also been used in other tasks such as visual explanation \citep{fukui2019attention}, video captioning \citep{yan2019stat}, healthcare \citep{choi2016retain}, object detection \citep{chorowski2015attention}, and speech recognition \citep{li2020object}.
The transformer architecture introduced in \citep{vaswani2017attention} is an autoencoder that encodes the input data and then decodes them to the target domain. It does not use recurrent modulus and just uses self-attention mechanism. It is a breakthrough in NLP and is the base for many language models including bidirectional encoder representations from transformers, BERT, \citep{devlin2019bert}, generative pretrained transformer, GPT-3, \citep{brown2020language}, Text-to-Text Transfer Transformer, T5, \citep{JMLR:v21:20-074} and Google’s Meena \citep{adiwardana2020towards}. It has also been successfully used as a baseline in other tasks such as object detection \citep{carion2020end}, image generation \citep{chen2021pre}, image colorization \citep{kumar2021colorization}, video understanding \citep{sun2019videobert}, and visual question answering \citep{tan2019lxmert}. Furthermore, \citet{yun2019transformers} showed that transformers can universally approximate sequence to sequence functions. Therefore, the transformer is a good choice for transfer learning not only because of their prosperity across different tasks, but also because of its architecture which makes it possible to use the hardware parallelism to train much more big models with much more training data.
\subsection{Symbolic computation related works}
The research on algebraic manipulation systems through computer is quite mature. The early work of solving symbolic integration were the heuristics programs written in LISP. They were named SIN (Symbolic INtegrator), SAINT, and SOlDIER (SOLUtion of Ordinary Differential Equations ROUTINE) \citep{moses1967symbolic}. The obvious motivation during those programs, is the use of symbolic systems as an adjunct to numerical integration programs which involves parameters. SAINT program of symbolic integration shows the capability of a freshman calculus student. Thus, an unmodified SAINT was of limited use in a practical algebraic system. More powerful programs follow, e.g., MATLAB project by MITRE Corporation, which solves integration of rational functions as good as sophomore college students. Though the capabilities of these programs are quite impressive, they mainly use tree search and matching algebraic expressions (pattern matching) as their workhorse. The program started showing its inherent limitation for those expressions which are not integrable in closed form, e.g., $\int e^{x^{2}} dx $ or $\int \frac{e^{x}}{x} dx$. Though there were some attempts of using Edge heuristics to solve those wild integrals, they were mainly unsuccessful. The era of deep neural networks ushers a new hope of solving the symbolic tasks by representing (encoding) the algebraic expressions in a feature space~\citep{Symbolic, wu2022autoformalization, arabshahi2018towards, allamanis2017learning, zaremba2014learning, loos2017deep, trask2018neural, kaiser2016neural, zaremba2015learning, valipour2021symbolicgpt, ling2017program, polu2020generative, hendrycksmath2021, lewkowycz2022solving, https://doi.org/10.48550/arxiv.2212.09196, wei2022emergent, doi:10.1073/pnas.2123433119, article}. 

Therefore, instead of pattern matching on the raw mathematical expressions done in the pre-deep learning era programs, these deep models solve the algebraic systems in the feature space. These works on representing the symbolic expressions in a continuous and differential space using deep net architectures show the fundamental difference in the philosophy from the early SIN, SAINT, and SOlDIER programs. The advantages of using deep net architectures are remarkable in terms of solving the algebraic systems approximately, e.g., for those integrals which have no closed form solutions, and the average time complexity to solve. The deep models even started to show creativity on solving complex mathematical expressions, e.g., representing a mathematical expression in multiple ways. Very recently, the research community started using language base transformer neural networks to solve symbolic computations~\citep{Symbolic, hendrycks2021measuring, https://doi.org/10.48550/arxiv.2112.01898, https://doi.org/10.48550/arxiv.2211.08671}. The mathematical expressions are encoded as a sequence and a transformer is trained for a sequence-to-sequence translation task. The dot product attention module in the transformer architecture solves symbolic tasks efficiently. 
\citet{saxton2019analysing} takes a different route and created a large symbolic mathematics data set. All these research directions point towards the direction of solving mathematics is no more in the genre of human creativity, but a data problem. The unreasonable effectiveness of symbolic mathematics data and large neural architectures show the inevitable future of machine generated mathematical solvers and symbolic mathematics.

\section{Conclusion}
\label{conclution}
Considering success of the transformer architecture in many tasks \citep{lu2021pretrained}, including both language and symbolic mathematics, we proposed transfer learning from a pretrained language model with the transformer architecture for the downstream task of solving symbolic mathematical problems such as integration and differential equations. Using multiple experimental evaluation, we showed that these models could achieve competitive performance (specially in the integration tasks) with transformers fully trained on the symbolic math task without being pretrained on linguistic data. We showed that the language that the transformer model has been pretrained on does not have a significant impact in this transfer learning. We also evaluated that a model fine-tuned using our approach generalizes better in distribution shift scenarios for integration tasks. 

\subsubsection*{Acknowledgments}
This work has been supported in part by NSF (Awards 2007202, 2107463, and 2038080), and the IT Academy Research Programme, Estonia. We are grateful to all who provided feedback on this work, including the anonymous reviewers of the International Conference on Learning Representations (ICLR).
\bibliography{refs}

\end{document}